\pdfoutput=1
\documentclass[11pt]{article}

\usepackage{acl}

\usepackage{times}
\usepackage{latexsym}

\usepackage[T1]{fontenc}
\usepackage[utf8]{inputenc}

\usepackage{microtype}

\usepackage{inconsolata}

\usepackage{url}
\usepackage{graphicx}
\usepackage{wrapfig}
\usepackage{amsmath}
\usepackage{amssymb}
\usepackage{color,xcolor,colortbl}
\usepackage{enumitem}
\usepackage{algorithm}
\usepackage{algorithmic}
\usepackage[font={small}]{caption}
\usepackage{bm,bbm}
\usepackage{booktabs}
\usepackage{mathtools}
\usepackage{array}
\usepackage{multirow}
\usepackage{soul}
\usepackage{pbox}
\usepackage{pifont}
\usepackage{arydshln}
\usepackage{comment}
\usepackage{array}
\usepackage{subcaption}

\newcommand\numberthis{\addtocounter{equation}{1}\tag{\theequation}}

\title{\textsc{Rescue}: Ranking LLM Responses with Partial Ordering to Improve \\Response Generation}

\author{Yikun Wang,$^1$ Rui Zheng,$^1$ Haoming Li,$^2$ Qi Zhang,$^1\thanks{~~Corresponding Author}$\; Tao Gui,$^1$ Fei Liu$^2$ \\[0.5em]
$^1$School of Computer Science, Fudan University\\
$^2$Computer Science Department, Emory University\\
\texttt{\{yikunwang19, rzheng20, qz, tgui\}@fudan.edu.cn}\\
\texttt{\{haoming.li, fei.liu\}@emory.edu}\\
}

\begin{document}
\maketitle

\begin{abstract}

Customizing LLMs for a specific task involves separating high-quality responses from  lower-quality ones. This skill can be developed using supervised fine-tuning with extensive human preference data. However, obtaining a large volume of expert-annotated data is costly for most tasks. In this paper, we explore a novel method to optimize LLMs using ranking metrics. This method trains the model to prioritize the best responses from a pool of candidates created for a particular task. Rather than a traditional full ordering, we advocate for a partial ordering, as achieving consensus on the perfect order of candidate responses can be challenging. Our partial ordering is more robust, less sensitive to noise, and can be achieved with limited human annotations or through heuristic methods. We test our system's improved response generation ability using benchmark datasets, including textual entailment and multi-document question answering. We conduct ablation studies to understand crucial factors, such as how to gather candidate responses for a specific task, determine their most suitable order, and balance supervised fine-tuning with ranking metrics. Our approach, named \textsc{Rescue}, offers a promising avenue for enhancing the response generation and task accuracy of LLMs.\footnote{Our code and models are available at: \url{https://github.com/ekonwang/RRescue}.}

\end{abstract}

\section{Introduction}
\label{sec:intro}

\begin{figure}
\centering
\includegraphics[width=2.95in]{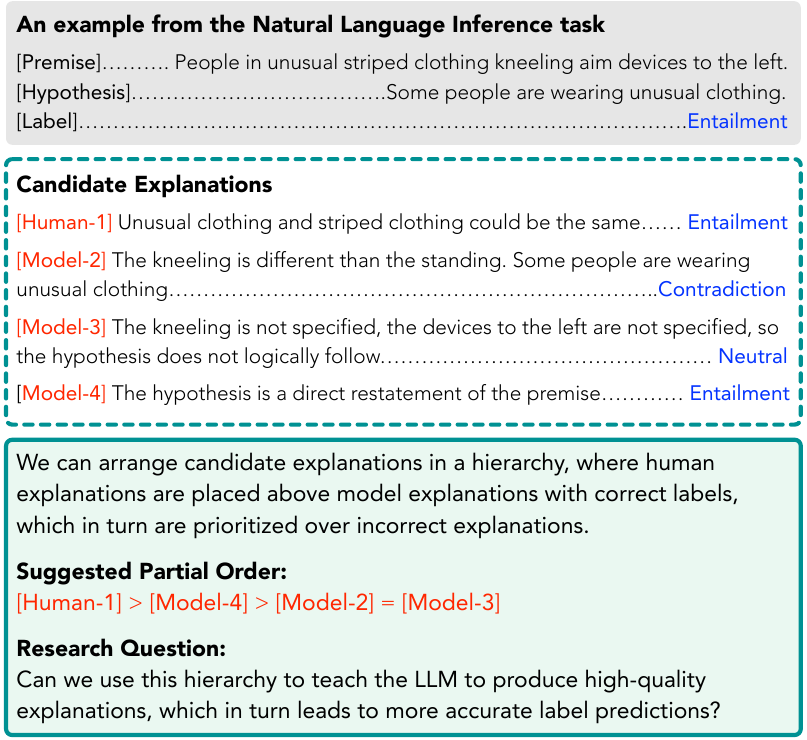}
\vspace{-0.05in}
\caption{When LLMs provide accurate label predictions, they are frequently accompanied by high-quality explanations~\cite{liu2023prudent}. Building on this insight, we rank candidate explanations obtained from diverse sources into a partial order. Human responses are placed above model responses with correct labels, and these are prioritized over incorrect responses. In scenarios with limited human annotations, we use this hierarchy to teach the LLM to generate high-quality explanations, which in turn leads to more accurate label predictions.}
\label{fig:example-snli}
\vspace{-0.2in}
\end{figure}

A significant advantage of large language models is their ability to explain their predictions \cite{ziegler2020finetuning,vafa-etal-2021-rationales,Alkhamissi_2023,ludan2023explanationbased,li2023explaining,ye2023complementary}. For example, LLMs may suggest lab tests to doctors based on patient symptoms~\cite{peng2023study} or help financial analysts evaluate risks in their investment portfolios~\cite{romanko2023chatgptbased}, providing explanations for each. As LLMs increasingly assist in decision-making across domains, examining the quality of their explanations becomes crucial. Previous studies suggest that lower-quality model explanations can lead to misunderstandings and diminish user trust~\cite{Burns:2022aa,Turpin:2023aa,reingold2024dissenting}. Therefore, it is imperative to improve LLMs' explanation quality, along with enhancing their task accuracy.

We focus on LLM responses that consist of \emph{a predicted label} and \emph{a detailed explanation}. LLMs should provide not only accurate labels but also sound rationales to support their predictions. Explanations can be generated using methods such as chain- or tree-of-thoughts and self-reflection~\cite{paper:yao2022react,wei2023chainofthought,paper:tree-of-thought,paper:reflexion,asai2023selfrag}. Explanations can also be embedded in prompts to guide LLMs in new tasks via in-context learning~\cite{ye2023complementary}. In this paper, we advance the research by investigating methods to train an open-source LLM to effectively rank candidate responses, which we gather from various sources. Learning to rank responses allows the LLM to differentiate between sound and flawed explanations for a specific task, thereby enhancing response generation.

Interestingly, accurate model predictions often come with high-quality explanations. Studies have shown that when LLMs are confident in their responses, they not only provide accurate answers but also offer solid justifications. On the flip side, when they're uncertain, their explanations can falter or be completely hallucinated~\cite{singh2023confidencecompetence,liu2023prudent,sun2024trustllm}. Our paper builds on this insight to rank candidate responses. We place human responses above model responses with correct labels, which in turn are prioritized over incorrect responses. This hierarchy encourages the LLM to generate explanations comparable to humans' or, at the very least, to produce explanations that lead to accurate labels.

Our method benefits from requiring minimal expert annotations, which is a frequent challenge in most domain-specific tasks. Unlike reinforcement learning with human feedback (RLHF; Ziegler et al., 2020)\nocite{ziegler2020finetuning} or direct preference optimization (DPO; Rafailov et al., 2023)\nocite{rafailov2023direct}, which need extensive expert-annotated data, our approach is cost-effective and practical in resource-constrained situations. We employ a partial ordering of LLM responses, which can be acquired with limited human annotations or through heuristic functions. This study's contributions are summarized as follows:
\begin{itemize}[topsep=5pt,itemsep=0pt,leftmargin=*]

\item We seek to improve LLMs' response generation. In training, we supplement each example with candidate responses, featuring a mix of accurate and inaccurate predictions, and sound and flawed explanations. For tasks with long contexts, we anchor responses in different parts of the context to increase diversity. LLM is trained to prioritize the best responses using the ranking metric. 

\item We test our system's response generation using multiple benchmarks, and conduct ablation studies to understand crucial factors, such as how to gather candidate responses, determine their most suitable order, and balance supervised fine-tuning with ranking metrics. Our approach, named \textsc{Rescue}, offers a promising avenue for enhancing the response generation and task accuracy of LLMs.

\end{itemize}

\section{Related Work}
\label{sec:related}

\paragraph{Learning from Human Preferences} Aligning LLM responses with human preferences ensures the models' outputs are helpful, safe, and adhere to societal norms \cite{paper:helpful-and-harmless,liu2023chain,honovich-etal-2023-unnatural,wang-etal-2023-self-instruct,rafailov2023direct,hejna2023contrastive,hu-etal-2023-decipherpref}. This research often involves humans performing pairwise or k-wise comparisons on model outputs, which are used to train a reward model~\cite{bai2022training,ouyang2022training,ramamurthy2023reinforcement,zhu2023principled}. 
Moreover, Rafailov et al.~\shortcite{rafailov2023direct} optimize the LLM directly based on preference data, eliminating the need for a separate reward model. 
Liu et al.~\shortcite{liu2024statistical} collect preference data from the target optimal policy through rejection sampling.
Unlike other methods, we guide LLMs to make accurate predictions and generate reliable explanations with minimal human annotations for domain-specific tasks.

\vspace{-0.05in}
\paragraph{Reasoning} LLMs can improve their reasoning through trial and error and self-improvement~\cite{wei2023chainofthought,Burnell:2023aa,zheng2023step,hu2024sportsmetrics,hu2024reasoning,Cheng:2024aa,Ahn:2024aa,Wang:2024aa}. For example, chain-of-thought \cite{wei2023chainofthought} allows LLMs to break down complex tasks step by step into more manageable parts. Tree-of-thoughts \cite{paper:tree-of-thought} employs task decomposition via a tree structure, guiding LLMs through various steps and consider multiple thoughts within each step. Reflexion \cite{paper:reflexion} combines dynamic memory and self-reflection to refine reasoning skills. However, pinpointing specific reasoning errors remains a practical challenge. The distinction between sound and flawed explanations can often be subtle and unclear during self-reflection.

\vspace{-0.05in}
\paragraph{Ranking Metrics} \quad A ranking objective allows the model to prioritize the best candidates~\cite{yuan2023rrhf,song2024preference}, improving its performance in tasks like abstractive summarization and question answering. For example, the BRIO training paradigm \cite{paper:brio} fine-tunes BART and T5 models to generate reference summaries while using a ranking mechanism to score candidate summaries. This approach could be especially beneficial in retrieval augmented generation \cite{hopkins-may-2011-tuning,lewis2021retrievalaugmented,nakano2022webgpt,hou2024large}. We believe that explanations grounded on incorrect documents should be discounted and those grounded in reference documents be promoted. Our method leverages this insight to enhance the model's ability to generate contextually accurate explanations.

\section{Our Approach: \textsc{Rescue}}
\label{sec:approach}

Let $x\sim\mathcal{D}$ represent the prompt or context given to the model, and $y$ denote the model's response to prompt $x$. The response $y$ comprises two parts: a brief justification and a predicted label, separated by the special symbol `\#\#\#\#'. For example, in the natural language inference task, it might be ``\emph{Unusual clothing and striped clothing could be the same. \#\#\#\# Entailment}.'' Supervised fine-tuning (SFT; Eq.~(\ref{eq:sft})) is a primary method to improve task accuracy by training the model to generate human-written responses $y^*$. However, since the model has only been exposed to high-quality human responses, its noise robustness remains unvalidated. Prior studies \cite{ziegler2020finetuning,paper:llama2-paper} suggest that model performance can plateau quickly, potentially leading to overfitting.
\begin{align*}
& \mathcal{L}_{\mbox{\scriptsize SFT}}(\theta) = - \log \pi_\theta(y^*|x) 
\numberthis\label{eq:sft}
\end{align*}

We proposed to guide the model to prioritize valid responses over flawed ones and contextually accurate responses over inaccurately grounded ones, using a ranking metric as illustrated in Eq.~(\ref{eq:rank}). Here, $(x,y_0,y_1,b)\sim\mathcal{S}$ includes a prompt $x$, two candidate responses, and a binary variable $b$, where $y_b$ should be scored higher than $y_{1-b}$. $\mathcal{S}$ represents a diverse set of candidate responses obtained from various sources. For example, responses could be acquired from open-source LLMs like Llama-2/3 or close-source LLMs like GPT-3.5, GPT-4 or Claude. Human-annotated responses can also be included in the collection when they are available. 
\begin{align*}
& \mathcal{L}_{\mbox{\scriptsize Rank}}(\theta) = - \mathbb{E}_{(x,y_0,y_1,b)\sim\mathcal{S}} [\ \numberthis\label{eq:rank} \\
& \quad\quad\quad \max\{0, \log \pi_\theta(y_b|x) - \log \pi_\theta(y_{1-b}|x)\}\ ]
\end{align*}

We initiate $\pi_\theta(y|x)$ from a base model $\rho(y|x)$ and subsequently fine-tune it for a specific task with candidate responses. Particularly, $\pi_\theta(y|x)$ is used to loosely represent length-normalized probability $\pi_\theta(y|x) = \frac{1}{|y|^\lambda}\sum_{t=1}^{|y|} \log \pi_\theta(y_t|x,y_{<t})$, where $\lambda > 0$ is the scaling factor for length normalization. Following Yuan et al.~\shortcite{yuan2023rrhf}, our approach uses $\alpha$ to balance the impact of supervised fine-tuning and the ranking metric, as shown in Eq.~(\ref{eq:rrescue}).
\begin{align*}
& \mathcal{L}_{\mbox{\scriptsize \textsc{Rescue}}}(\theta) = \mathcal{L}_{\mbox{\scriptsize SFT}}(\theta) + \alpha \mathcal{L}_{\mbox{\scriptsize Rank}}(\theta)\numberthis\label{eq:rrescue}
\end{align*}

\paragraph{Ranking Metrics vs. Rewards} A reward model $r(x,y_i)$ assigns scores to a given prompt $x$ and its corresponding response $y_i$. As shown in Eq.~(\ref{eq:reward}), it allocates the \emph{full probability mass} to the response $y_b$ chosen by human labelers. For this model to function, humans need to provide accurate pairwise preference judgments. Nonetheless, achieving a consensus among human labelers regarding the perfect order of LLM responses can be a daunting task. The labelers often struggle to provide consistent, fine-grained labels~\cite{paper:llama2-paper}. As a result, allocating the entire probability mass, i.e., $\log \mathcal{P}_\theta(y_{b'}|x)$ to an incorrectly labeled response $y_{b'}$ can mislead the model and hinder the effective training of the reward model.
\begin{align*}
 \mathcal{L}_{\mbox{\scriptsize Reward}}(r) = & 
- \mathbb{E}_{(x,\{y_i\}_i,b)\sim\mathcal{S}} \Bigg[\log \frac{e^{r(x,y_b)}}{\sum_i e^{r(x,y_i)}} \Bigg]
\numberthis\label{eq:reward}
\end{align*}

In contrast, our proposed ranking metrics offer greater flexibility and robustness to inconsistencies in human preferences. Our model not only prioritizes $y_b$ over other potential responses using the equation $\max\{0, \log \mathcal{P}_\theta(y_b|x) - \log \mathcal{P}_\theta(y_{1-b}|x)\}$, but further allows minor deviations. For example, the model can still assign a high probability to a less-favored response $\log \mathcal{P}_\theta(y_{1-b}|x)$, provided its probability difference from the top response $\log \mathcal{P}_\theta(y_b|x) - \log \mathcal{P}_\theta(y_{1-b}|x)$ remains minimal. We also advocate for a partial ordering of LLM responses, partitioning them into groups. This group ordering provides a hierarchical perspective, enabling the model to understand the relative importance of each group in a broader context.

\section{Ranking LLM Responses}
\label{sec:organizing_res}

Candidate responses for a given prompt $x$, can be organized into a strict order. OpenAI has employed a team of trained human labelers to rank sets of model outputs from best to worst to train a reward model~\cite{ziegler2020finetuning,ouyang2022training}. However, this method is quite expensive. We propose two cost-effective approaches to establish a Partial Ordering (PO) of responses.

Our first method, \textbf{(PO) Human Prioritization}, posits that human responses should take priority over model responses, as they offer valid rationales and accurate labels. \textbf{(PO) Label Prioritization} places responses with correct labels above those with incorrect labels, irrespective of whether they are human or model-generated. This is because rationales resulting in correct labels are more valuable than those leading to incorrect labels. The latter may contain flawed reasoning that misguides their predictions. Lastly, \textbf{(PO) Human-Label Hybrid} employs a fine-grained grouping. It places human responses above model responses with correct labels, which are then prioritized over responses with incorrect labels. This hierarchy is designed to motivate the LLM to generate rationales comparable to humans' or, at a minimum, to produce rationales that lead to accurate labels.

Partial Orderings (PO) of responses offer enhanced flexibility and noise robustness. For example, in developing Llama-2, Touvron et al. \shortcite{paper:llama2-paper} noted that even human labelers struggle to decide between two similar model responses, with annotations for such responses often hinging on subjective judgement and nuanced details. By utilizing a partial order, we only incorporate the most clear-cut pairs of model outputs in the ranking metric, thereby improving the quality of response pairs used in model fine-tuning.

For comparison, we examine two full ordering (FO) approaches. \textbf{(FO) Similarity} embeds each candidate response into a vector, which are then ranked based on their Cosine similarity to the vector representing the human response. The second approach \textbf{(FO) GPT-3.5-Turbo} leverages the GPT-3.5-Turbo-0613 model to rank candidate responses. We instruct it to prioritize candidates with the same labels as the human response, but allowing it to decide whether this criterion is met. We compare full and partial ordering approaches in \S\ref{sec:experiments}.

\section{Collecting Candidate Responses}
\label{sec:cand_responses}

We enrich each example with a set of candidate responses, targeting a mix that includes both accurate and inaccurate predictions, along with explanations that are both sound and flawed. We incorporate human annotations into the mix when available. For tasks with long contexts, we anchor responses in different parts of the context to increase diversity. This enriched dataset is used to train our LLM to improve its response generation. Next, we outline two strategies for generating candidate responses.

\subsection{Responses Generated by Various LLMs}
\label{sec:cand-responses-snli}

We focus on the textual entailment task~\cite{bowman-etal-2015-large,chen-etal-2017-enhanced,paper:camburu2018esnli,kumar-talukdar-2020-nile} to illustrate our strategy. Specifically, the Stanford NLI dataset identifies relationships between sentence pairs as \emph{entailment}, \emph{contradiction}, or \emph{neutral}. The e-SNLI dataset expands on SNLI by adding human-annotated explanations for these relationships, explaining why sentences are classified in certain ways~\cite{paper:camburu2018esnli}. Similarly, we require LLMs to both \emph{predict and rationalize} their predictions. Our approach then learns to prioritize accurate predictions and their model explanations, while downplaying explanations for inaccurate predictions.

We gather diverse responses for this task from both open-source and proprietary LLMs. Specifically, we sample three responses from \texttt{Llama-2-7b}, setting the temperature to 0.8 for diversity, and one from \texttt{GPT-3.5-Turbo-0613}, plus a human explanation, making five responses per prompt in total. Each response features a brief explanation of the model's reasoning and a predicted label, as shown in Figure~\ref{fig:example-snli}.

\paragraph{Response Flipping} We propose a novel method for collecting diverse responses from LLMs without the need for repetitive response sampling. Our method begins by inverting an LLM's explanation for a given response. For instance, if an LLM suggests, ``\emph{The to-go packages may not be from lunch. \#\#\#\# Neutral,}'' we flip the explanation to, ``\emph{The to-go packages are likely from lunch.}'' This reversed explanation then guides the LLM to assign a new label, such as ``\emph{\#\#\#\# Entailment.}''

Our method uses \texttt{GPT-4-0613} for reversing the explanations, given its extraordinary generation capabilities. The prompt for inversion is: ``\emph{Rewrite the sentence to convey the opposite meaning: \{Explanation\}}.'' Afterward, \texttt{GPT-3.5-Turbo-0613} is used to predict the appropriate label by combining the original context with the inverted explanation. This method offers an efficient way to generate diverse responses with varying labels.

\begin{figure}
\centering
\includegraphics[width=3in]{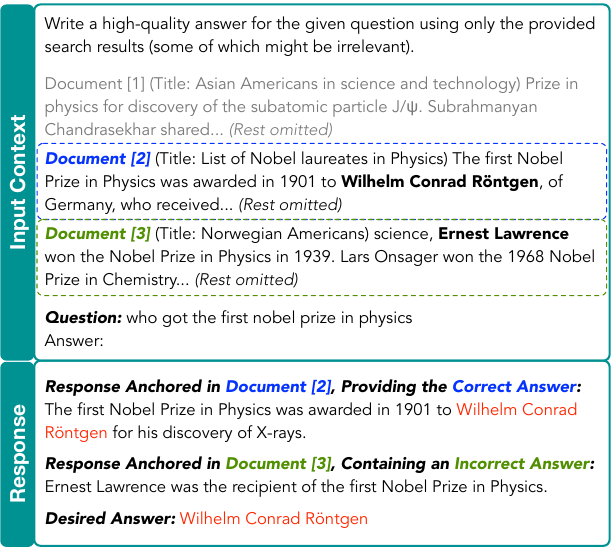}
\vspace{-0.15in}
\caption{
For the Multi-doc QA task, we anchor responses in different parts of the context to produce a diverse set of answers. We generate five candidate responses per instance, one from the gold passage and four from random distractors. }
\label{fig:task-qa}
\vspace{-0.1in}
\end{figure}

\subsection{Responses Anchored in Various Passages}
\label{sec:cand-responses-multidoc}

When dealing with long contexts, we can anchor responses in different parts of the context to produce a diverse set of answers. An LLM can then enhance its performance by discriminating among these answers. For example, in the multi-document question answering task (\textbf{Multi-doc QA}; Liu et al. 2023b)\nocite{paper:nelson_liu_lost}, the LLM uses 10 to 30 Wikipedia passages as input to answer questions. These questions come from NaturalQuestions-Open~\cite{kwiatkowski-etal-2019-natural}, which contains historical Google queries and their human-annotated answers extracted from Wikipedia. Among the passages given to the model, only one has the answer, the rest are distractors. A retrieval system named Contriever~\cite{izacard2022unsupervised} is used to obtain distractor passages, which are most relevant to the question but do not contain the answers. 

We use \texttt{Llama-2-7b} to generate five diverse candidate responses per instance, one from the gold passage and four from random distractors. Responses containing the desired answer are marked correct, as illustrated in Figure~\ref{fig:task-qa}. Here, we generate two candidate responses ``\emph{The first Nobel Prize in Physics was awarded in 1901 to Wilhelm Conrad Röntgen for his discovery of X-rays.}'' and ``\emph{Ernest Lawrence was the recipient of the first Nobel Prize in Physics.}'' by feeding the model Documents [2] and [3] separately. Our Label-Prioritized approach ranks candidates with the desired answer higher than those without. Human-Label-Hybrid further prefers correct answers anchored in the gold passage. In training, the model receives a question and 10 Wikipedia passages, and learns to differentiate correct from incorrect responses. At test time, the fine-tuned model employs beam search to decode the optimal response.

\begin{table*}[!t]
\setlength{\tabcolsep}{5.5pt}
\renewcommand{\arraystretch}{1}
\centering
\begin{tabular}{llccccccc} 
\toprule
& & \multicolumn{5}{c}{\textbf{Proportion of Training Data}} & \multicolumn{2}{c}{\textbf{w/ Res. Flip.}} \\
& \textbf{System} & 0.4\% & 0.9\% & 1.8\% & 3.6\% & \textsc{Avg} & 0.4\% & 0.9\% \\ 
\midrule
\textsc{Baseline} & (SFT) Supervised Finetuning & 77.45 & 85.56 & 87.33 & 87.94 & 84.57 &  -- & --\\
& (FO) Similarity & 81.01 & 86.69 & 86.53 & 86.38 & 85.15 & $\boldsymbol\uparrow$ 5.18 & $\boldsymbol\downarrow$ 0.26 \\
& (FO) GPT-3.5-Turbo & 82.20 & 86.62 & 85.02 & 86.71 & 85.14 & $\boldsymbol\uparrow$ 3.09 & $\boldsymbol\downarrow$ 1.32 \\
\cmidrule{2-9}
\textsc{Ours} & (PO) Human Prioritization & 80.70 & 87.11 & 87.06 & 86.26 & 85.28 & $\boldsymbol\uparrow$ 6.10 & $\boldsymbol\downarrow$ 1.30 \\
& \textcolor{red}{(PO) Label Prioritization} & 81.97 & 87.27 & \textbf{88.16} & \textbf{87.97} & 86.34 & $\boldsymbol\uparrow$ 5.15 & $\boldsymbol\uparrow$ 0.61 \\
& \textcolor{red}{(PO) Human-Label Hybrid} & \textbf{82.86} & \textbf{87.47} & 87.33 & 87.73 & \textcolor{red}{\textbf{86.35}} & $\boldsymbol\uparrow$ 4.88 & $\boldsymbol\uparrow$ 0.34 \\
\bottomrule
\end{tabular}
\caption{
Task accuracy of \textsc{Rescue} on natural language inference, reported on the e-SNLI test set. 
We observe that models trained with ranking metrics and incorporating both full and partial ordering strategies outperform those trained solely with SFT, especially when working with a few thousand annotated examples. 
Our partial ordering strategies, namely label prioritization and a hybrid of human and label prioritization, surpass full ordering methods.
}
\label{table:main}
\vspace{-0.1in}
\end{table*}

\section{Experiments}
\label{sec:experiments}

We have chosen \texttt{Llama-2-7b} as our base model for task-specific training. The \texttt{Llama-2} series outperforms other open-source options, such as Falcon~\cite{almazrouei2023falcon}, Mistral~\cite{jiang2023mistral}, Vicuna~\cite{vicuna2023} and MPT~\cite{paper:MPT-30b}, on a number of tasks. Its \texttt{7b} variant requires significantly less GPU memory, which is crucial for specific domains without the specialized infrastructure to serve larger models.\footnote{We leave the extension to other models such as Llama-3 for future work.}

We use AdamW~\cite{paper:adamw} with a learning rate of $2e^{-5}$ and a cosine scheduler with a $0.03$ warmup rate. Our training utilizes fully sharded data parallelism and BF16 mixed precision training, which is generally faster, consumes less memory, and is preferable for large models. Our experiments are conducted using 4xA100 GPUs, and task-specific training is limited to a single epoch for both supervised fine-tuning and response ranking. This is to mitigate the risk of multi-epoch degradation \cite{paper:multiepoch-degradation} and potential overfitting from repeated exposure to the training data. 
The batch size is set at $B$=64, the same configuration used for \texttt{LLama-2}~\cite{paper:llama2-paper}. It is calculated as the product of three factors, $B = g \times b \times D$, combining gradient accumulation steps ($g=16$), per-GPU batch size ($b=1$ due to memory constraints), and the number of GPUs ($D=4$). This strategy allows us to handle a large number of candidates during response ranking.

\begin{figure}
\centering
\includegraphics[width=3in]{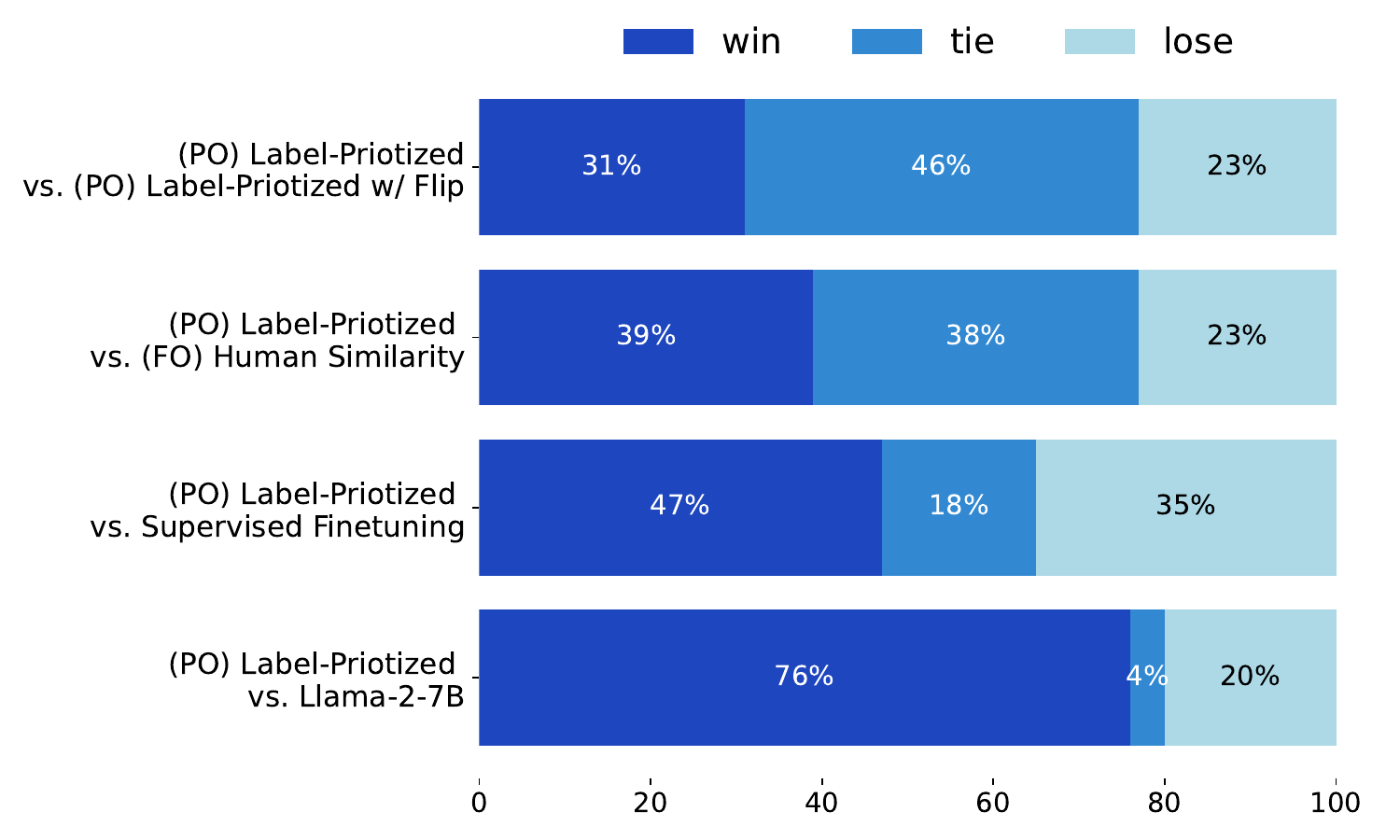}
\vspace{-0.2in}
\caption{
Human evaluation results.
Our partial ordering (PO) with label prioritization outperforms the SFT model with an overall win rate of 47\%. While SFT shows comparable accuracy in automatic evaluation, it often relies on data artifacts for predictions~\cite{DBLP:journals/corr/abs-1803-02324} and does not yield better explanations. Our PO method also outperforms other methods such as FO Similarity and the base \texttt{Llama-2-7b} model. 
}  
\vspace{-0.1in}
\label{fig:winrate}
\end{figure}

\subsection{Automatic Evaluation of NLI Accuracy}
\label{sec:results-snli}

Our goal in this study is to enhance response generation with limited training data, which is a common challenge in real-world scenarios where expert annotations are scarce, often limited to a few thousand examples. We conduct our experiments using the e-SNLI dataset~\cite{paper:camburu2018esnli}, which comprises 549,367 training examples. We intentionally restrict our training to subsets of \{2k, 5k, 10k, 20k\} samples, approximately 0.4\% to 3.6\% of the total training set. We report the accuracy of all models on the standard test set of 9,824 examples.

We evaluate a variety of models on this task. In particular, we train the base model with human responses (\textbf{SFT}). We also explore two response ranking strategies: full ordering ({FO}), which ranks candidate model responses by their semantic closeness to human responses (\textbf{Similarity}) or as assessed by \textbf{GPT-3.5-Turbo}, and partial ordering ({PO}), which trains the base model to prioritize human responses over those from models (\textbf{Human Prioritization}), responses with correct labels over incorrect ones (\textbf{Label Prioritization}), and a mix of both (\textbf{Human-Label Hybrid}). Both FO and PO rely on our ranking metric detailed in Eq.(\ref{eq:rrescue}).

Table~\ref{table:main} presents task accuracy across various proportions of training data. We observe that models trained with ranking metrics and incorporating both full and partial ordering strategies outperform those trained solely with SFT, especially when working with a few thousand annotated examples. This indicates that training an LLM to rank responses can improve response generation and result in more accurate predictions of textual entailment relationships. The improvement is most notable when using only 0.4\% of the total training data, suggesting the advantage of ranking metrics in scenarios with extremely scarce training data.

\begin{table*}[t]
\setlength{\tabcolsep}{6.8pt}
\renewcommand{\arraystretch}{1}
\centering
\begin{tabular}{lcccccccc} 
\toprule
& \multicolumn{4}{c}{{\textbf{5 Retrieved Documents}}} & \multicolumn{4}{c}{{\textbf{10 Retrieved Documents}}} \\
\textbf{Position of Gold Document} & 1st & 3rd & 5th & \textsc{Avg} & 1st & 5th & 10th & \textsc{Avg} \\
\midrule
Base Model (\texttt{Llama-2-7b})  & \textbf{45.64} & 34.19 & 43.05 & 40.96 & \textbf{46.41} & 27.17 & 42.95 & 38.84 \\
\textcolor{red}{(PO) Label Prioritization} & 44.88 & \textbf{42.44} & \textbf{53.43} & \textcolor{red}{\textbf{46.92}} & 35.72 & \textbf{33.43} & \textbf{55.11} & \textcolor{red}{\textbf{41.42}} \\
\bottomrule
\end{tabular}
\vspace{-0.05in}
\caption{
Answer accuracy for the Multi-QA task. We evaluate two scenarios: the model receives 5 or 10 documents returned by the retriever. We find that the PO method with label prioritization substantially improves model performance, as ranking responses allows the LLM to more effectively identify relevant information, improving the U-shaped curve. 
}
\label{table:mdoc}
\vspace{-0.1in}
\end{table*}

\begin{figure*}
\centering
\includegraphics[width=0.4\linewidth]{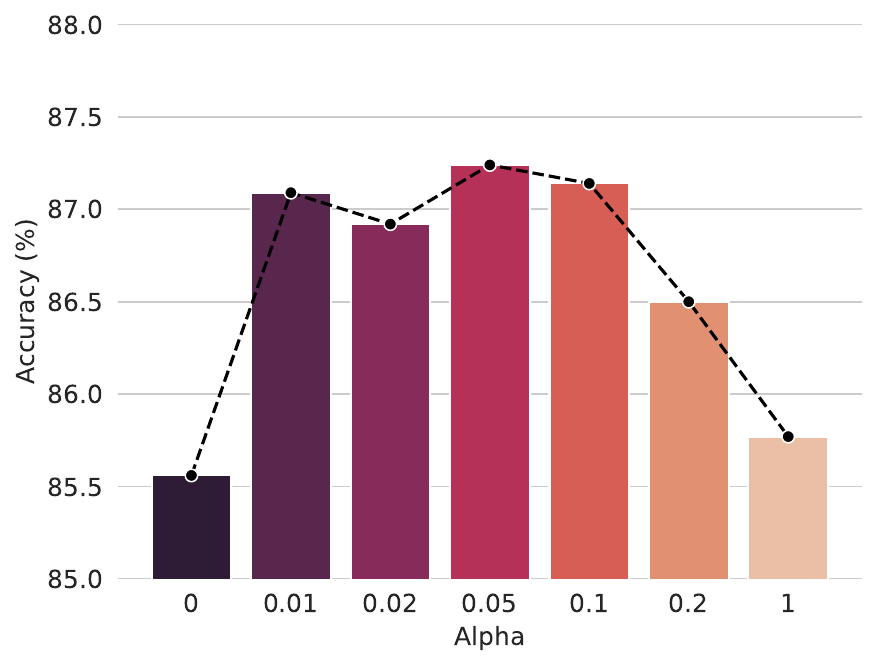}
\quad\quad\quad
\includegraphics[width=0.4\linewidth]{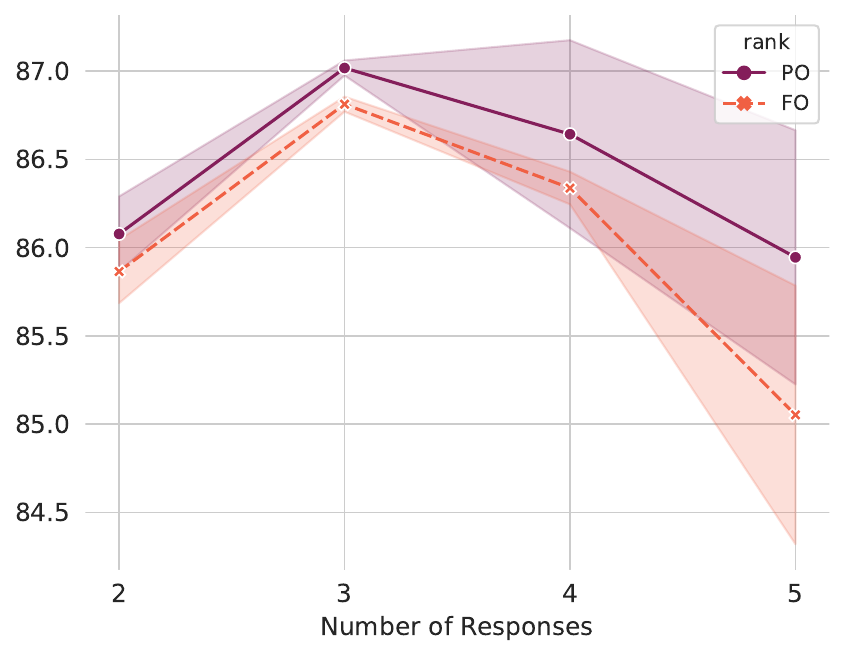}
\vspace{-0.05in}
\caption{
(\textsc{Left}) The influence of different $\alpha$ on task accuracy. We find that optimal performance is achieved with an $\alpha$ value between 0.01 to 0.1. (\textsc{Right}) We conduct experiments with a varying number of candidate responses per prompt. Results indicate that performance improvement can be achieved even with 3-4 candidate responses. 
}  
\label{fig:hyperparameter}
\vspace{-0.1in}
\end{figure*}

Our partial ordering strategies, namely label prioritization and a hybrid of human and label prioritization, surpass full ordering methods. This could be because achieving consensus on full ordering of responses is challenging even for humans. This approach may introduce variability in response ranking and destabilizes training. SFT begins to show improvement with 20k or more training examples, although gathering such extensive annotations is often difficult for domain-specific tasks. Additionally, while flipping responses increases answer variety, it might cause a shift in the distribution of ranked responses. We find this technique consistently improves response generation only when training data is limited to 2k examples. 

Our models match state-of-the-art performance. E.g., Hsieh et al.~\shortcite{paper:DSBS} achieved 89.51\% accuracy using a 540B LLM with step-by-step distilling. By contrast, our models use only a fraction of the full training set with a 7B model. Without supervised fine-tuning, the base \texttt{Llama-2-7b} model yields a significantly lower accuracy of 33.31\%. Next, we extend our evaluation to include human assessment of model explanations.

\begin{figure*}
    \centering
    \begin{subfigure}{.29\textwidth}
    \includegraphics[width=\linewidth]{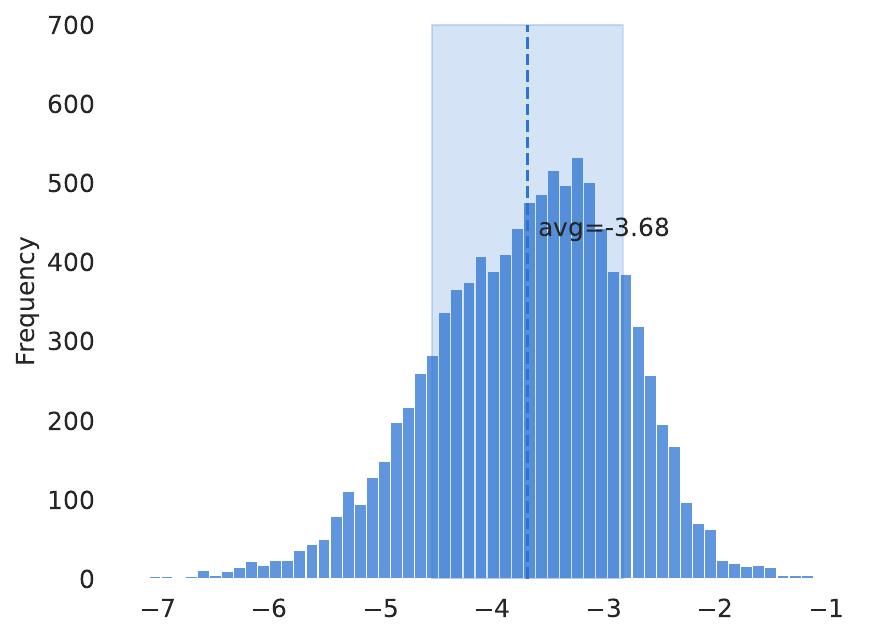}
    \label{fig:dist-score-human}
    \end{subfigure}
    \begin{subfigure}{.32\textwidth}
    \includegraphics[width=\linewidth]{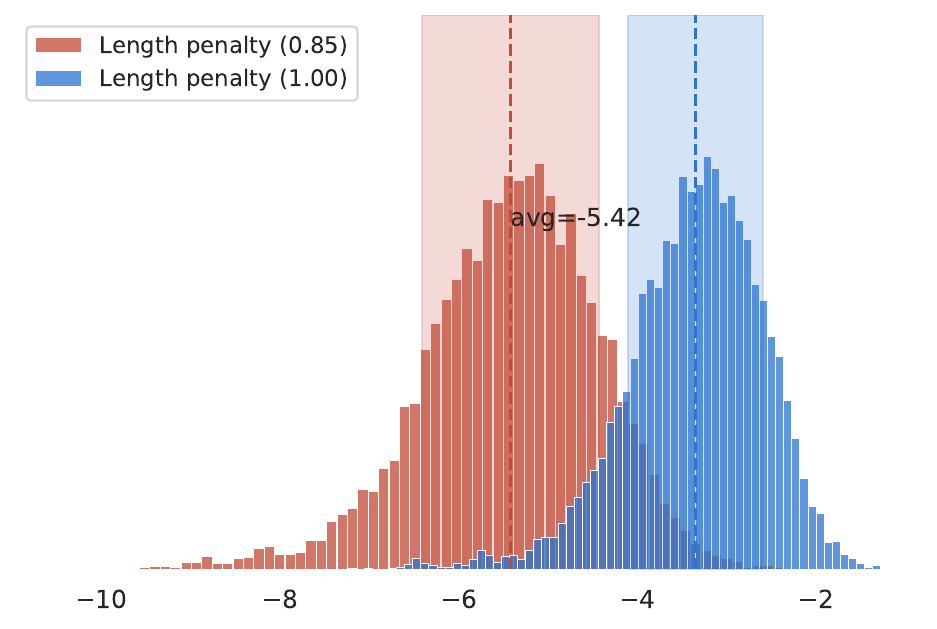}
    \label{fig:dist-score-llama2-lp-0.85}
    \end{subfigure}
    \begin{subfigure}{.32\textwidth}
    \includegraphics[width=\linewidth]{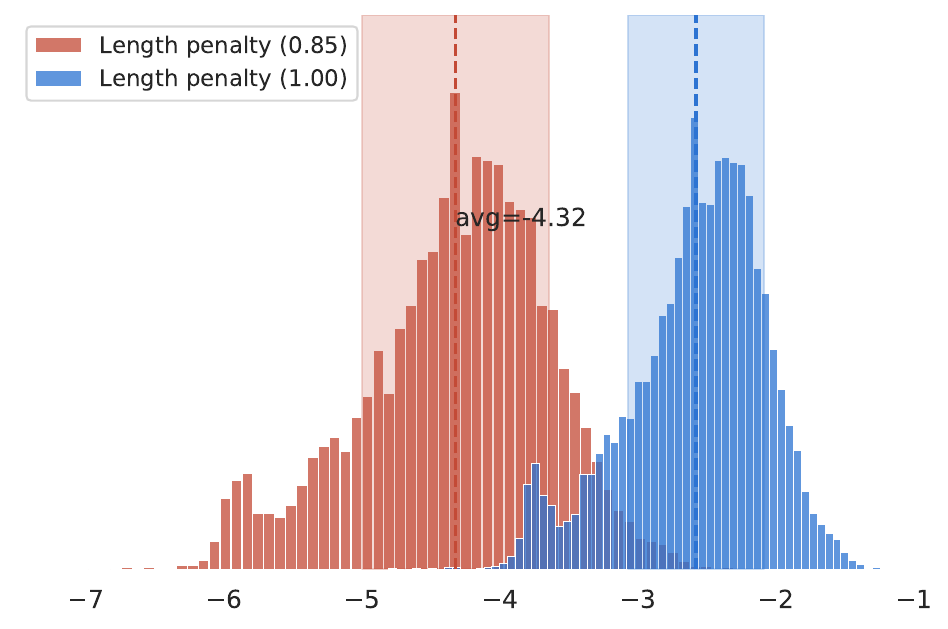}
    \label{fig:dist-score-gpt-3.5-0.85}
    \end{subfigure}
\vspace{-0.15in}
\caption{
\textsc{Left} figure shows the log probabilities of human responses, while \textsc{Middle} and \textsc{Right} figures present those from Llama-2-7b and GPT-3.5-turbo-0613, respectively. We assign a length scaling factor, $\lambda$, of 0.85 to all model responses, maintaining a $\lambda$ of 1.0 for human responses. This approach effectively shifts the log probability score distributions of model responses (colored in red) closer to those of human ones, thereby minimizing margin violations.
}
\label{fig:dist-scores}
\vspace{-0.1in}
\end{figure*}

\subsection{Human Evaluation of Response Quality} 
\label{sec:human_eval}

Human evaluation provides a holistic assessment of model responses. We compare several models, including our PO method with label prioritization, SFT, FO method with responses ranked their similarity to human responses, PO model with response flipping, and the base model. These models were trained with varying amounts of training data (0.4\% to 3.6\%), and the highest performing model across all data proportions was chosen for human evaluation. An annotator evaluated responses for 100 randomly selected samples from the e-SNLI test set, using win, tie and lose to rate each response pair. Evaluations were based on label accuracy and the quality of explanations. A quality explanation should support the predict label with detailed reasoning and show logical coherence.

As Figure~\ref{fig:winrate} illustrates, our partial ordering (PO) with label prioritization outperforms the SFT model with an overall win rate of 47\%. This advantage stems from the PO models' ability to distinguish between sound and flawed responses, thus improving response generation. While SFT shows comparable accuracy in automatic evaluation, it often relies on data artifacts for predictions~\cite{DBLP:journals/corr/abs-1803-02324} and does not yield better explanations. Similar to findings from automatic evaluations, adding response flipping does not surpass the original label prioritization method. Our PO method also outperforms other methods such as FO Similarity and the base \texttt{Llama-2-7b} model.

\subsection{Evaluation of Multi-Document QA}
\label{sec:results-multidoc}

The Multi-Doc QA task involves answering a given question using a set of retrieved documents. Liu et al.~\shortcite{paper:nelson_liu_lost} found that LLMs exhibit a U-shaped  curve, depending on where the answer-containing document is located within the input context and highlighting difficulties in accessing relevant information in the middle of long contexts. To mitigate this, we incorporate response ranking. We generate five candidate responses per question, one from the correct document and four from distractors. We then train the base model on 1k examples from the training set using our ranking metric (Eq. (\ref{eq:rank})). SFT is not used due to the absence of human-written explanations for this task. Our method is evaluated on a test set of 665 examples. 

Table~\ref{table:mdoc} shows answer accuracy, measured as whether correct answers from the NaturalQuestions annotations appear in the generated responses. We evaluate two scenarios: the model receives 5 or 10 documents returned by the retriever. The correct document is placed either at the beginning (1st position), in the middle (3rd or 5th), or at the end (5th or 10th) of the document set. We find that the PO method with label prioritization substantially improves model performance, as ranking responses allows the LLM to more effectively identify relevant information, improving the U-shaped curve. Our findings also align with those of Liu et al.~\shortcite{paper:nelson_liu_lost}, who observed a recency bias in Llama-2-7b. With 20 documents as input, they reported accuracies of about 25\% at positions 1, 5, 10, 15, and 42\% at position 20. Upon examining the model's responses, we observe that the model often answers questions by copying content, which tends to improve answer accuracy when the answer is located in the middle or end of the context.

\begin{figure*}
\centering
\includegraphics[width=0.45\linewidth]{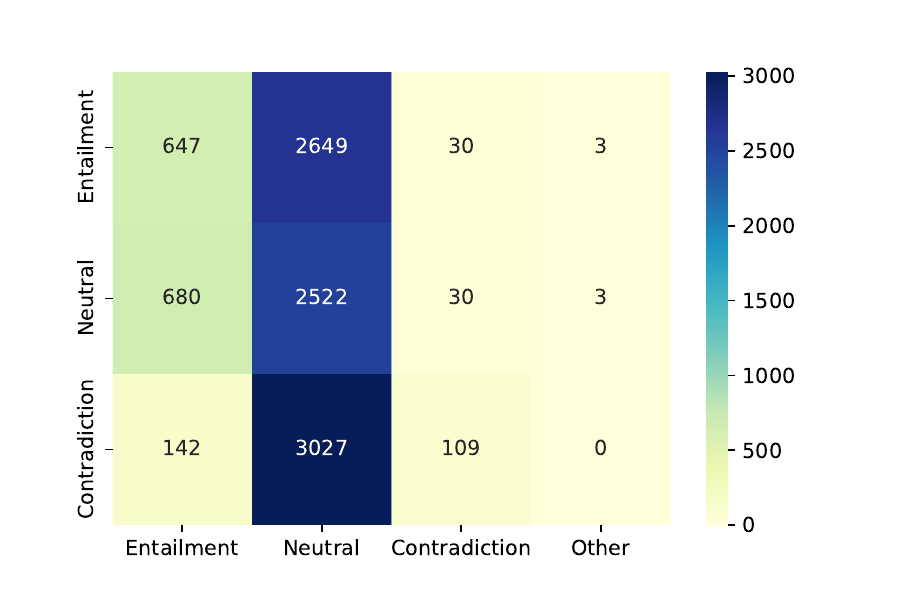}\hfill
\includegraphics[width=0.53\linewidth]{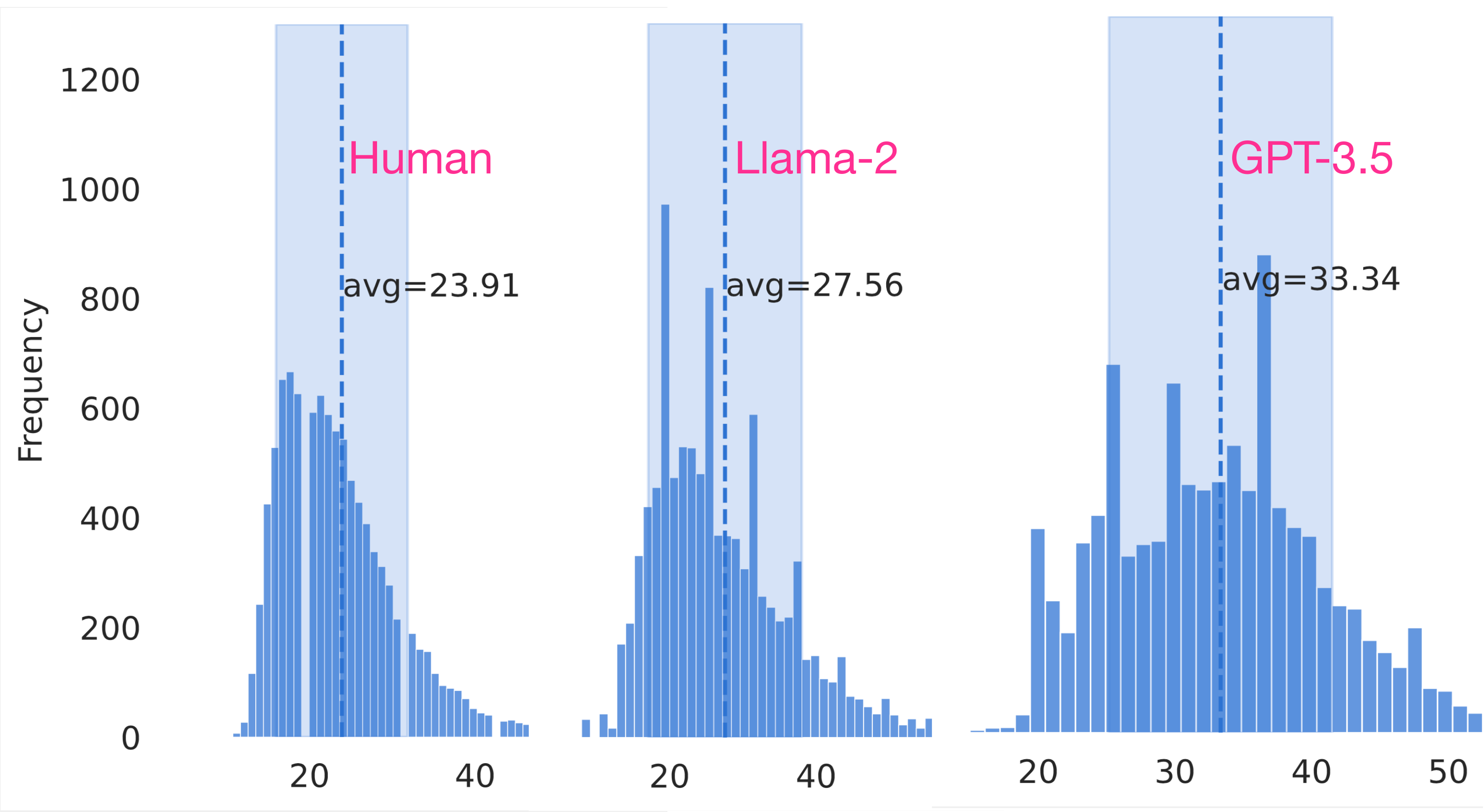}
\caption{
(\textsc{Left}) The confusion matrix for the Llama-2-7b base model, where the x-axis represents the labels predicted by Llama-2-7b, and the y-axis represents human labels. The results show Llama-2-7b's tendency to predict neutral labels, as indicated by the dark bar in the middle.
(\textsc{Right}) Candidate responses differ in length. We show the distribution of responses from human annotators, Llama-2-7b, and GPT-3.5-turbo-0613 models. Human responses are the shortest, while GPT-3.5's are notably longer, containing on average 10 more tokens per response compared to human responses.
}  
\label{fig:cm}
\vspace{-0.1in}
\end{figure*}

\section{Discussion}
\label{subsection:discussion}

\paragraph{Balancing Coefficient} Our approach uses a hyperparameter $\alpha$ to balance the impact of supervised fine-tuning and the ranking metric. Figure \ref{fig:hyperparameter} shows the influence of different $\alpha$ on task accuracy. We find that optimal performance is achieved with an $\alpha$ value between 0.01 to 0.1. Our results indicate that, while supervised fine-tuning is pivotal for \textsc{Rescue}, integrating the ranking metric enhances the method's robustness to noise.

\paragraph{Number of Candidate Responses} We conduct experiments with a varying number of candidate responses per prompt, and the results are shown in Figure \ref{fig:hyperparameter}. In our experiments, we are able to rank up to five candidate responses using four Nvidia A100 GPUs. As the number of candidates increases, so does the demand for additional GPU memory and compute resources. Our experiments indicate that performance improvement can be achieved even with 3-4 candidate responses. Beyond that, \textsc{Rescue} sees no further gains from increasing the number of responses. This saturation in performance may be attributed to the noise in ranking. Moreover, it highlights the challenges associated with ranking a diverse set of responses differing in length and style of explanations.

\paragraph{Scoring Candidate Responses} We identify two characteristics in human responses that distinguish them from model responses. Firstly, they are more concise and to the point. As indicated in Figure~\ref{fig:cm} (\textsc{Right}), human responses are significantly shorter, averaging 10 fewer tokens per response compared to GPT-3.5's responses. Secondly, we note that LLM responses tend to use more common words, yielding better fluency and generally smoother text compared to human responses. These characteristics present challenges in ranking responses from diverse sources. Human responses, due to their brevity and unique word choice, often have lower length-normalized log probabilities than model responses. This discrepancy leads to many margin violations during training using Eq. (\ref{eq:rank}), and more parameter updates to ensure human responses score higher than model outputs.

To mitigate this, we assign a length scaling factor $\lambda$ of 0.85 to all model responses, including those from Llama-2-7b and GPT-3.5-turbo-0613, maintaining a $\lambda$ of 1.0 for human responses. This effectively shifts the log probability score distributions for model responses closer to human ones (Figure \ref{fig:dist-scores}), reducing margin violations. We are also exploring adjusting the margin size and curriculum learning, which gradually increases the difficulty of training samples to reduce violations, as potential directions for future research.

\paragraph{Central Tendency Bias} LLMs such as Llama-2-7b and GPT-3.5 exhibit a central tendency bias~\cite{goldfarb-tarrant-etal-2020-content} in natural language inference. These models often predict \emph{Neutral} labels, leaning towards the ``center'' of possible labels. Figure \ref{fig:cm} presents the confusion matrix, with the x-axis representing predicted labels by Llama-2-7b and the y-axis showing human labels. The results show Llama-2-7b's tendency to predict neutral labels (indicated by the dark bar in the middle) and its avoidance of extreme labels like \emph{Entailment} or \emph{Contradiction}. A plausible reason could be Llama-2-7b's inadequate world knowledge impacting its task accuracy. Moreover, this tendency might originate from the models being trained on human annotations for instruction-following. They frequently give hedging responses to fulfill helpfulness and safety requirements, leading to outputs that are more neutral and less assertive.

\vspace{-0.05in}
\section{Conclusion}
\label{sec:conclusion}

In this paper, we introduce \textsc{Rescue}, an approach that trains the LLM to prioritize sound responses over erroneous ones, thereby enhancing overall task accuracy and the quality of explanations. Accurate model predictions often come with high-quality explanations. We build on this insight to rank candidate responses using a partial ordering approach, as achieving consensus on the perfect order of responses is challenging. \textsc{Rescue} has demonstrated competitive performance on benchmarks.

\section*{Acknowledgements}

We would like to thank the reviewers for their insightful feedback, which greatly enhanced our paper. HL and FL are supported in part by National Science Foundation grant IIS-2303678.

\section*{Limitations}
\label{sec:limitations}

Our approach focuses on optimizing LLMs through ranking metrics and partial ordering of candidate responses. We introduce two innovative strategies for generating candidates: collecting from diverse LLMs and anchoring responses in various parts of the context, showcasing its flexibility across benchmark datasets. We note that organizing candidate responses can benefit from domain-specific criteria, such as sorting recommended lab tests for patients by the relevance of the answer, urgency, and cost. Further, our proposed approach prioritizes the best responses from a set of candidates, thereby improving the task accuracy and the quality of generated explanations. With additional GPU resources, we can improve the variety and representation of candidate responses or categorize them based on domain-specific attributes. Despite existing challenges, our approach offers a promising path for customizing LLMs for specialized applications.

\bibliography{anthology,custom}

\begin{thebibliography}{62}
\expandafter\ifx\csname natexlab\endcsname\relax\def\natexlab#1{#1}\fi

\bibitem[{Ahn et~al.(2024)Ahn, Verma, Lou, Liu, Zhang, and Yin}]{Ahn:2024aa}
Janice Ahn, Rishu Verma, Renze Lou, Di~Liu, Rui Zhang, and Wenpeng Yin. 2024.
\newblock \href {http://arxiv.org/abs/2402.00157} {Large language models for
  mathematical reasoning: Progresses and challenges}.

\bibitem[{Alkhamissi et~al.(2023)Alkhamissi, Verma, Yu, Jin, Celikyilmaz, and
  Diab}]{Alkhamissi_2023}
Badr Alkhamissi, Siddharth Verma, Ping Yu, Zhijing Jin, Asli Celikyilmaz, and
  Mona Diab. 2023.
\newblock \href {https://doi.org/10.18653/v1/2023.nlrse-1.10} {{OPT-R}:
  Exploring the role of explanations in finetuning and prompting for reasoning
  skills of large language models}.
\newblock In \emph{Proceedings of the 1st Workshop on Natural Language
  Reasoning and Structured Explanations (NLRSE)}. Association for Computational
  Linguistics.

\bibitem[{Almazrouei et~al.(2023)Almazrouei, Alobeidli, Alshamsi, Cappelli,
  Cojocaru, Debbah, Étienne Goffinet, Hesslow, Launay, Malartic, Mazzotta,
  Noune, Pannier, and Penedo}]{almazrouei2023falcon}
Ebtesam Almazrouei, Hamza Alobeidli, Abdulaziz Alshamsi, Alessandro Cappelli,
  Ruxandra Cojocaru, Mérouane Debbah, Étienne Goffinet, Daniel Hesslow,
  Julien Launay, Quentin Malartic, Daniele Mazzotta, Badreddine Noune, Baptiste
  Pannier, and Guilherme Penedo. 2023.
\newblock \href {http://arxiv.org/abs/2311.16867} {The falcon series of open
  language models}.

\bibitem[{Asai et~al.(2024)Asai, Wu, Wang, Sil, and
  Hajishirzi}]{asai2023selfrag}
Akari Asai, Zeqiu Wu, Yizhong Wang, Avirup Sil, and Hannaneh Hajishirzi. 2024.
\newblock {Self-RAG}: Learning to retrieve, generate, and critique through
  self-reflection.
\newblock In \emph{Proceedings of the 2024 International Conference on Learning
  Representations (ICLR)}.

\bibitem[{Bai et~al.(2022{\natexlab{a}})Bai, Jones, Ndousse, Askell, Chen,
  DasSarma, Drain, Fort, Ganguli, Henighan, Joseph, Kadavath, Kernion, Conerly,
  El-Showk, Elhage, Hatfield-Dodds, Hernandez, Hume, Johnston, Kravec, Lovitt,
  Nanda, Olsson, Amodei, Brown, Clark, McCandlish, Olah, Mann, and
  Kaplan}]{bai2022training}
Yuntao Bai, Andy Jones, Kamal Ndousse, Amanda Askell, Anna Chen, Nova DasSarma,
  Dawn Drain, Stanislav Fort, Deep Ganguli, Tom Henighan, Nicholas Joseph,
  Saurav Kadavath, Jackson Kernion, Tom Conerly, Sheer El-Showk, Nelson Elhage,
  Zac Hatfield-Dodds, Danny Hernandez, Tristan Hume, Scott Johnston, Shauna
  Kravec, Liane Lovitt, Neel Nanda, Catherine Olsson, Dario Amodei, Tom Brown,
  Jack Clark, Sam McCandlish, Chris Olah, Ben Mann, and Jared Kaplan.
  2022{\natexlab{a}}.
\newblock Training a helpful and harmless assistant with reinforcement learning
  from human feedback.
\newblock \emph{arXiv preprint 2204.05862}.

\bibitem[{Bai et~al.(2022{\natexlab{b}})Bai, Jones, Ndousse, Askell, Chen,
  DasSarma, Drain, Fort, Ganguli, Henighan et~al.}]{paper:helpful-and-harmless}
Yuntao Bai, Andy Jones, Kamal Ndousse, Amanda Askell, Anna Chen, Nova DasSarma,
  Dawn Drain, Stanislav Fort, Deep Ganguli, Tom Henighan, et~al.
  2022{\natexlab{b}}.
\newblock Training a helpful and harmless assistant with reinforcement learning
  from human feedback.
\newblock \emph{arXiv preprint arXiv:2204.05862}.

\bibitem[{Bowman et~al.(2015)Bowman, Angeli, Potts, and
  Manning}]{bowman-etal-2015-large}
Samuel~R. Bowman, Gabor Angeli, Christopher Potts, and Christopher~D. Manning.
  2015.
\newblock \href {https://doi.org/10.18653/v1/D15-1075} {A large annotated
  corpus for learning natural language inference}.
\newblock In \emph{Proceedings of the 2015 Conference on Empirical Methods in
  Natural Language Processing}, pages 632--642, Lisbon, Portugal. Association
  for Computational Linguistics.

\bibitem[{Burnell et~al.(2023)Burnell, Hao, Conway, and
  Orallo}]{Burnell:2023aa}
Ryan Burnell, Han Hao, Andrew R.~A. Conway, and Jose~Hernandez Orallo. 2023.
\newblock \href {http://arxiv.org/abs/2306.10062} {Revealing the structure of
  language model capabilities}.

\bibitem[{Burns et~al.(2022)Burns, Ye, Klein, and Steinhardt}]{Burns:2022aa}
Collin Burns, Haotian Ye, Dan Klein, and Jacob Steinhardt. 2022.
\newblock \href {http://arxiv.org/abs/2212.03827} {Discovering latent knowledge
  in language models without supervision}.

\bibitem[{Camburu et~al.(2018)Camburu, Rockt{\"{a}}schel, Lukasiewicz, and
  Blunsom}]{paper:camburu2018esnli}
Oana{-}Maria Camburu, Tim Rockt{\"{a}}schel, Thomas Lukasiewicz, and Phil
  Blunsom. 2018.
\newblock e-snli: Natural language inference with natural language
  explanations.
\newblock In \emph{Advances in Neural Information Processing Systems 31: Annual
  Conference on Neural Information Processing Systems 2018, NeurIPS 2018,
  December 3-8, 2018, Montr{\'{e}}al, Canada}, pages 9560--9572.

\bibitem[{Chen et~al.(2017)Chen, Zhu, Ling, Wei, Jiang, and
  Inkpen}]{chen-etal-2017-enhanced}
Qian Chen, Xiaodan Zhu, Zhen-Hua Ling, Si~Wei, Hui Jiang, and Diana Inkpen.
  2017.
\newblock \href {https://doi.org/10.18653/v1/P17-1152} {Enhanced {LSTM} for
  natural language inference}.
\newblock In \emph{Proceedings of the 55th Annual Meeting of the Association
  for Computational Linguistics (Volume 1: Long Papers)}, pages 1657--1668,
  Vancouver, Canada. Association for Computational Linguistics.

\bibitem[{Cheng et~al.(2024)Cheng, Ahmed, Willke, and Sun}]{Cheng:2024aa}
Kewei Cheng, Nesreen~K. Ahmed, Theodore Willke, and Yizhou Sun. 2024.
\newblock \href {http://arxiv.org/abs/2402.13415} {Structure guided prompt:
  Instructing large language model in multi-step reasoning by exploring graph
  structure of the text}.

\bibitem[{Chiang et~al.(2023)Chiang, Li, Lin, Sheng, Wu, Zhang, Zheng, Zhuang,
  Zhuang, Gonzalez, Stoica, and Xing}]{vicuna2023}
Wei-Lin Chiang, Zhuohan Li, Zi~Lin, Ying Sheng, Zhanghao Wu, Hao Zhang, Lianmin
  Zheng, Siyuan Zhuang, Yonghao Zhuang, Joseph~E. Gonzalez, Ion Stoica, and
  Eric~P. Xing. 2023.
\newblock \href {https://lmsys.org/blog/2023-03-30-vicuna/} {Vicuna: An
  open-source chatbot impressing gpt-4 with 90\%* chatgpt quality}.

\bibitem[{Goldfarb-Tarrant et~al.(2020)Goldfarb-Tarrant, Chakrabarty,
  Weischedel, and Peng}]{goldfarb-tarrant-etal-2020-content}
Seraphina Goldfarb-Tarrant, Tuhin Chakrabarty, Ralph Weischedel, and Nanyun
  Peng. 2020.
\newblock \href {https://doi.org/10.18653/v1/2020.emnlp-main.351} {Content
  planning for neural story generation with aristotelian rescoring}.
\newblock In \emph{Proceedings of the 2020 Conference on Empirical Methods in
  Natural Language Processing (EMNLP)}, pages 4319--4338, Online. Association
  for Computational Linguistics.

\bibitem[{Gururangan et~al.(2018)Gururangan, Swayamdipta, Levy, Schwartz,
  Bowman, and Smith}]{DBLP:journals/corr/abs-1803-02324}
Suchin Gururangan, Swabha Swayamdipta, Omer Levy, Roy Schwartz, Samuel~R.
  Bowman, and Noah~A. Smith. 2018.
\newblock \href {http://arxiv.org/abs/1803.02324} {Annotation artifacts in
  natural language inference data}.
\newblock \emph{CoRR}, abs/1803.02324.

\bibitem[{Hejna et~al.(2023)Hejna, Rafailov, Sikchi, Finn, Niekum, Knox, and
  Sadigh}]{hejna2023contrastive}
Joey Hejna, Rafael Rafailov, Harshit Sikchi, Chelsea Finn, Scott Niekum,
  W.~Bradley Knox, and Dorsa Sadigh. 2023.
\newblock \href {http://arxiv.org/abs/2310.13639} {Contrastive preference
  learning: Learning from human feedback without rl}.

\bibitem[{Honovich et~al.(2023)Honovich, Scialom, Levy, and
  Schick}]{honovich-etal-2023-unnatural}
Or~Honovich, Thomas Scialom, Omer Levy, and Timo Schick. 2023.
\newblock \href {https://aclanthology.org/2023.acl-long.806} {Unnatural
  instructions: Tuning language models with (almost) no human labor}.
\newblock In \emph{Proceedings of the 61st Annual Meeting of the Association
  for Computational Linguistics (Volume 1: Long Papers)}, pages 14409--14428,
  Toronto, Canada. Association for Computational Linguistics.

\bibitem[{Hopkins and May(2011)}]{hopkins-may-2011-tuning}
Mark Hopkins and Jonathan May. 2011.
\newblock \href {https://aclanthology.org/D11-1125} {Tuning as ranking}.
\newblock In \emph{Proceedings of the 2011 Conference on Empirical Methods in
  Natural Language Processing}, pages 1352--1362, Edinburgh, Scotland, UK.
  Association for Computational Linguistics.

\bibitem[{Hou et~al.(2024)Hou, Zhang, Lin, Lu, Xie, McAuley, and
  Zhao}]{hou2024large}
Yupeng Hou, Junjie Zhang, Zihan Lin, Hongyu Lu, Ruobing Xie, Julian McAuley,
  and Wayne~Xin Zhao. 2024.
\newblock \href {http://arxiv.org/abs/2305.08845} {Large language models are
  zero-shot rankers for recommender systems}.

\bibitem[{Hsieh et~al.(2023)Hsieh, Li, Yeh, Nakhost, Fujii, Ratner, Krishna,
  Lee, and Pfister}]{paper:DSBS}
Cheng{-}Yu Hsieh, Chun{-}Liang Li, Chih{-}Kuan Yeh, Hootan Nakhost, Yasuhisa
  Fujii, Alex Ratner, Ranjay Krishna, Chen{-}Yu Lee, and Tomas Pfister. 2023.
\newblock \href {https://doi.org/10.18653/v1/2023.findings-acl.507} {Distilling
  step-by-step! outperforming larger language models with less training data
  and smaller model sizes}.
\newblock In \emph{Findings of the Association for Computational Linguistics:
  {ACL} 2023, Toronto, Canada, July 9-14, 2023}, pages 8003--8017. Association
  for Computational Linguistics.

\bibitem[{Hu et~al.(2023)Hu, Song, Cho, Wang, Foroosh, and
  Liu}]{hu-etal-2023-decipherpref}
Yebowen Hu, Kaiqiang Song, Sangwoo Cho, Xiaoyang Wang, Hassan Foroosh, and Fei
  Liu. 2023.
\newblock \href {https://doi.org/10.18653/v1/2023.emnlp-main.519}
  {{D}ecipher{P}ref: Analyzing influential factors in human preference
  judgments via {GPT}-4}.
\newblock In \emph{Proceedings of the 2023 Conference on Empirical Methods in
  Natural Language Processing}, pages 8344--8357, Singapore. Association for
  Computational Linguistics.

\bibitem[{Hu et~al.(2024{\natexlab{a}})Hu, Song, Cho, Wang, Foroosh, Yu, and
  Liu}]{hu2024sportsmetrics}
Yebowen Hu, Kaiqiang Song, Sangwoo Cho, Xiaoyang Wang, Hassan Foroosh, Dong Yu,
  and Fei Liu. 2024{\natexlab{a}}.
\newblock \href {http://arxiv.org/abs/2402.10979} {Sportsmetrics: Blending text
  and numerical data to understand information fusion in llms}.

\bibitem[{Hu et~al.(2024{\natexlab{b}})Hu, Song, Cho, Wang, Yao, Foroosh, Yu,
  and Liu}]{hu2024reasoning}
Yebowen Hu, Kaiqiang Song, Sangwoo Cho, Xiaoyang Wang, Wenlin Yao, Hassan
  Foroosh, Dong Yu, and Fei Liu. 2024{\natexlab{b}}.
\newblock \href {http://arxiv.org/abs/2406.12084} {When reasoning meets
  information aggregation: A case study with sports narratives}.

\bibitem[{Izacard et~al.(2022)Izacard, Caron, Hosseini, Riedel, Bojanowski,
  Joulin, and Grave}]{izacard2022unsupervised}
Gautier Izacard, Mathilde Caron, Lucas Hosseini, Sebastian Riedel, Piotr
  Bojanowski, Armand Joulin, and Edouard Grave. 2022.
\newblock \href {http://arxiv.org/abs/2112.09118} {Unsupervised dense
  information retrieval with contrastive learning}.

\bibitem[{Jiang et~al.(2023)Jiang, Sablayrolles, Mensch, Bamford, Chaplot,
  de~las Casas, Bressand, Lengyel, Lample, Saulnier, Lavaud, Lachaux, Stock,
  Scao, Lavril, Wang, Lacroix, and Sayed}]{jiang2023mistral}
Albert~Q. Jiang, Alexandre Sablayrolles, Arthur Mensch, Chris Bamford,
  Devendra~Singh Chaplot, Diego de~las Casas, Florian Bressand, Gianna Lengyel,
  Guillaume Lample, Lucile Saulnier, Lélio~Renard Lavaud, Marie-Anne Lachaux,
  Pierre Stock, Teven~Le Scao, Thibaut Lavril, Thomas Wang, Timothée Lacroix,
  and William~El Sayed. 2023.
\newblock \href {http://arxiv.org/abs/2310.06825} {Mistral 7b}.

\bibitem[{Kumar and Talukdar(2020)}]{kumar-talukdar-2020-nile}
Sawan Kumar and Partha Talukdar. 2020.
\newblock \href {https://doi.org/10.18653/v1/2020.acl-main.771} {{NILE} :
  Natural language inference with faithful natural language explanations}.
\newblock In \emph{Proceedings of the 58th Annual Meeting of the Association
  for Computational Linguistics}, pages 8730--8742, Online. Association for
  Computational Linguistics.

\bibitem[{Kwiatkowski et~al.(2019)Kwiatkowski, Palomaki, Redfield, Collins,
  Parikh, Alberti, Epstein, Polosukhin, Devlin, Lee, Toutanova, Jones, Kelcey,
  Chang, Dai, Uszkoreit, Le, and Petrov}]{kwiatkowski-etal-2019-natural}
Tom Kwiatkowski, Jennimaria Palomaki, Olivia Redfield, Michael Collins, Ankur
  Parikh, Chris Alberti, Danielle Epstein, Illia Polosukhin, Jacob Devlin,
  Kenton Lee, Kristina Toutanova, Llion Jones, Matthew Kelcey, Ming-Wei Chang,
  Andrew~M. Dai, Jakob Uszkoreit, Quoc Le, and Slav Petrov. 2019.
\newblock \href {https://doi.org/10.1162/tacl_a_00276} {Natural questions: A
  benchmark for question answering research}.
\newblock \emph{Transactions of the Association for Computational Linguistics},
  7:452--466.

\bibitem[{Lewis et~al.(2021)Lewis, Perez, Piktus, Petroni, Karpukhin, Goyal,
  Küttler, Lewis, tau Yih, Rocktäschel, Riedel, and
  Kiela}]{lewis2021retrievalaugmented}
Patrick Lewis, Ethan Perez, Aleksandra Piktus, Fabio Petroni, Vladimir
  Karpukhin, Naman Goyal, Heinrich Küttler, Mike Lewis, Wen tau Yih, Tim
  Rocktäschel, Sebastian Riedel, and Douwe Kiela. 2021.
\newblock \href {http://arxiv.org/abs/2005.11401} {Retrieval-augmented
  generation for knowledge-intensive nlp tasks}.

\bibitem[{Li et~al.(2023)Li, Tworkowski, Wu, and Mooney}]{li2023explaining}
Jierui Li, Szymon Tworkowski, Yingying Wu, and Raymond Mooney. 2023.
\newblock \href {http://arxiv.org/abs/2307.05337} {Explaining competitive-level
  programming solutions using llms}.

\bibitem[{Liu et~al.(2023{\natexlab{a}})Liu, Wang, Yuan, Chen, and
  Peng}]{liu2023prudent}
Genglin Liu, Xingyao Wang, Lifan Yuan, Yangyi Chen, and Hao Peng.
  2023{\natexlab{a}}.
\newblock \href {http://arxiv.org/abs/2311.09731} {Prudent silence or foolish
  babble? examining large language models' responses to the unknown}.

\bibitem[{Liu et~al.(2023{\natexlab{b}})Liu, Sferrazza, and
  Abbeel}]{liu2023chain}
Hao Liu, Carmelo Sferrazza, and Pieter Abbeel. 2023{\natexlab{b}}.
\newblock \href {http://arxiv.org/abs/2302.02676} {Chain of hindsight aligns
  language models with feedback}.

\bibitem[{Liu et~al.(2023{\natexlab{c}})Liu, Lin, Hewitt, Paranjape,
  Bevilacqua, Petroni, and Liang}]{paper:nelson_liu_lost}
Nelson~F. Liu, Kevin Lin, John Hewitt, Ashwin Paranjape, Michele Bevilacqua,
  Fabio Petroni, and Percy Liang. 2023{\natexlab{c}}.
\newblock \href {https://doi.org/10.48550/arXiv.2307.03172} {Lost in the
  middle: How language models use long contexts}.
\newblock \emph{CoRR}, abs/2307.03172.

\bibitem[{Liu et~al.(2024)Liu, Zhao, Joshi, Khalman, Saleh, Liu, and
  Liu}]{liu2024statistical}
Tianqi Liu, Yao Zhao, Rishabh Joshi, Misha Khalman, Mohammad Saleh, Peter~J.
  Liu, and Jialu Liu. 2024.
\newblock \href {http://arxiv.org/abs/2309.06657} {Statistical rejection
  sampling improves preference optimization}.

\bibitem[{Liu et~al.(2022)Liu, Liu, Radev, and Neubig}]{paper:brio}
Yixin Liu, Pengfei Liu, Dragomir Radev, and Graham Neubig. 2022.
\newblock \href {http://arxiv.org/abs/2203.16804} {Brio: Bringing order to
  abstractive summarization}.

\bibitem[{Loshchilov and Hutter(2017)}]{paper:adamw}
Ilya Loshchilov and Frank Hutter. 2017.
\newblock Decoupled weight decay regularization.
\newblock \emph{arXiv preprint arXiv:1711.05101}.

\bibitem[{Ludan et~al.(2023)Ludan, Meng, Nguyen, Shah, Lyu, Apidianaki, and
  Callison-Burch}]{ludan2023explanationbased}
Josh~Magnus Ludan, Yixuan Meng, Tai Nguyen, Saurabh Shah, Qing Lyu, Marianna
  Apidianaki, and Chris Callison-Burch. 2023.
\newblock \href {http://arxiv.org/abs/2305.04990} {Explanation-based finetuning
  makes models more robust to spurious cues}.

\bibitem[{MosaicML(2023)}]{paper:MPT-30b}
MosaicML. 2023.
\newblock Introducing mpt-30b: Raising the bar for open-source foundation
  models.
\newblock Accessed: 2023-06-22.

\bibitem[{Nakano et~al.(2022)Nakano, Hilton, Balaji, Wu, Ouyang, Kim, Hesse,
  Jain, Kosaraju, Saunders, Jiang, Cobbe, Eloundou, Krueger, Button, Knight,
  Chess, and Schulman}]{nakano2022webgpt}
Reiichiro Nakano, Jacob Hilton, Suchir Balaji, Jeff Wu, Long Ouyang, Christina
  Kim, Christopher Hesse, Shantanu Jain, Vineet Kosaraju, William Saunders,
  Xu~Jiang, Karl Cobbe, Tyna Eloundou, Gretchen Krueger, Kevin Button, Matthew
  Knight, Benjamin Chess, and John Schulman. 2022.
\newblock Webgpt: Browser-assisted question-answering with human feedback.
\newblock \emph{arXiv preprint 2112.09332}.

\bibitem[{Ouyang et~al.(2022)Ouyang, Wu, Jiang, Almeida, Wainwright, Mishkin,
  Zhang, Agarwal, Slama, Ray, Schulman, Hilton, Kelton, Miller, Simens, Askell,
  Welinder, Christiano, Leike, and Lowe}]{ouyang2022training}
Long Ouyang, Jeff Wu, Xu~Jiang, Diogo Almeida, Carroll~L. Wainwright, Pamela
  Mishkin, Chong Zhang, Sandhini Agarwal, Katarina Slama, Alex Ray, John
  Schulman, Jacob Hilton, Fraser Kelton, Luke Miller, Maddie Simens, Amanda
  Askell, Peter Welinder, Paul Christiano, Jan Leike, and Ryan Lowe. 2022.
\newblock \href {http://arxiv.org/abs/2203.02155} {Training language models to
  follow instructions with human feedback}.

\bibitem[{Peng et~al.(2023)Peng, Yang, Chen, Smith, PourNejatian, Costa,
  Martin, Flores, Zhang, Magoc, Lipori, Mitchell, Ospina, Ahmed, Hogan,
  Shenkman, Guo, Bian, and Wu}]{peng2023study}
Cheng Peng, Xi~Yang, Aokun Chen, Kaleb~E Smith, Nima PourNejatian, Anthony~B
  Costa, Cheryl Martin, Mona~G Flores, Ying Zhang, Tanja Magoc, Gloria Lipori,
  Duane~A Mitchell, Naykky~S Ospina, Mustafa~M Ahmed, William~R Hogan,
  Elizabeth~A Shenkman, Yi~Guo, Jiang Bian, and Yonghui Wu. 2023.
\newblock \href {http://arxiv.org/abs/2305.13523} {A study of generative large
  language model for medical research and healthcare}.

\bibitem[{Rafailov et~al.(2023)Rafailov, Sharma, Mitchell, Ermon, Manning, and
  Finn}]{rafailov2023direct}
Rafael Rafailov, Archit Sharma, Eric Mitchell, Stefano Ermon, Christopher~D.
  Manning, and Chelsea Finn. 2023.
\newblock \href {http://arxiv.org/abs/2305.18290} {Direct preference
  optimization: Your language model is secretly a reward model}.

\bibitem[{Ramamurthy et~al.(2023)Ramamurthy, Ammanabrolu, Brantley, Hessel,
  Sifa, Bauckhage, Hajishirzi, and Choi}]{ramamurthy2023reinforcement}
Rajkumar Ramamurthy, Prithviraj Ammanabrolu, Kianté Brantley, Jack Hessel,
  Rafet Sifa, Christian Bauckhage, Hannaneh Hajishirzi, and Yejin Choi. 2023.
\newblock Is reinforcement learning (not) for natural language processing:
  Benchmarks, baselines, and building blocks for natural language policy
  optimization.
\newblock \emph{arXiv preprint 2210.01241}.

\bibitem[{Reingold et~al.(2024)Reingold, Shen, and
  Talati}]{reingold2024dissenting}
Omer Reingold, Judy~Hanwen Shen, and Aditi Talati. 2024.
\newblock \href {http://arxiv.org/abs/2307.07636} {Dissenting explanations:
  Leveraging disagreement to reduce model overreliance}.

\bibitem[{Romanko et~al.(2023)Romanko, Narayan, and
  Kwon}]{romanko2023chatgptbased}
Oleksandr Romanko, Akhilesh Narayan, and Roy~H. Kwon. 2023.
\newblock \href {http://arxiv.org/abs/2308.06260} {Chatgpt-based investment
  portfolio selection}.

\bibitem[{Shinn et~al.(2023)Shinn, Cassano, Labash, Gopinath, Narasimhan, and
  Yao}]{paper:reflexion}
Noah Shinn, Federico Cassano, Beck Labash, Ashwin Gopinath, Karthik Narasimhan,
  and Shunyu Yao. 2023.
\newblock \href {http://arxiv.org/abs/2303.11366} {Reflexion: Language agents
  with verbal reinforcement learning}.

\bibitem[{Singh et~al.(2023)Singh, Devkota, Lamichhane, Dhakal, and
  Dhakal}]{singh2023confidencecompetence}
Aniket~Kumar Singh, Suman Devkota, Bishal Lamichhane, Uttam Dhakal, and Chandra
  Dhakal. 2023.
\newblock \href {http://arxiv.org/abs/2309.16145} {The confidence-competence
  gap in large language models: A cognitive study}.

\bibitem[{Song et~al.(2024)Song, Yu, Li, Yu, Huang, Li, and
  Wang}]{song2024preference}
Feifan Song, Bowen Yu, Minghao Li, Haiyang Yu, Fei Huang, Yongbin Li, and
  Houfeng Wang. 2024.
\newblock \href {http://arxiv.org/abs/2306.17492} {Preference ranking
  optimization for human alignment}.

\bibitem[{Sun et~al.(2024)Sun, Huang, Wang, Wu, Zhang, Gao, Huang, Lyu, Zhang,
  Li, Liu, Liu, Wang, Zhang, Kailkhura, Xiong, Xiao, Li, Xing, Huang, Liu, Ji,
  Wang, Zhang, Yao, Kellis, Zitnik, Jiang, Bansal, Zou, Pei, Liu, Gao, Han,
  Zhao, Tang, Wang, Mitchell, Shu, Xu, Chang, He, Huang, Backes, Gong, Yu,
  Chen, Gu, Xu, Ying, Ji, Jana, Chen, Liu, Zhou, Wang, Li, Zhang, Wang, Xie,
  Chen, Wang, Liu, Ye, Cao, Chen, and Zhao}]{sun2024trustllm}
Lichao Sun, Yue Huang, Haoran Wang, Siyuan Wu, Qihui Zhang, Chujie Gao, Yixin
  Huang, Wenhan Lyu, Yixuan Zhang, Xiner Li, Zhengliang Liu, Yixin Liu, Yijue
  Wang, Zhikun Zhang, Bhavya Kailkhura, Caiming Xiong, Chaowei Xiao, Chunyuan
  Li, Eric Xing, Furong Huang, Hao Liu, Heng Ji, Hongyi Wang, Huan Zhang,
  Huaxiu Yao, Manolis Kellis, Marinka Zitnik, Meng Jiang, Mohit Bansal, James
  Zou, Jian Pei, Jian Liu, Jianfeng Gao, Jiawei Han, Jieyu Zhao, Jiliang Tang,
  Jindong Wang, John Mitchell, Kai Shu, Kaidi Xu, Kai-Wei Chang, Lifang He,
  Lifu Huang, Michael Backes, Neil~Zhenqiang Gong, Philip~S. Yu, Pin-Yu Chen,
  Quanquan Gu, Ran Xu, Rex Ying, Shuiwang Ji, Suman Jana, Tianlong Chen,
  Tianming Liu, Tianyi Zhou, William Wang, Xiang Li, Xiangliang Zhang, Xiao
  Wang, Xing Xie, Xun Chen, Xuyu Wang, Yan Liu, Yanfang Ye, Yinzhi Cao, Yong
  Chen, and Yue Zhao. 2024.
\newblock \href {http://arxiv.org/abs/2401.05561} {Trustllm: Trustworthiness in
  large language models}.

\bibitem[{Touvron et~al.(2023)Touvron, Martin, Stone, Albert, Almahairi,
  Babaei, Bashlykov, Batra, Bhargava, Bhosale et~al.}]{paper:llama2-paper}
Hugo Touvron, Louis Martin, Kevin Stone, Peter Albert, Amjad Almahairi, Yasmine
  Babaei, Nikolay Bashlykov, Soumya Batra, Prajjwal Bhargava, Shruti Bhosale,
  et~al. 2023.
\newblock Llama 2: Open foundation and fine-tuned chat models.
\newblock \emph{arXiv preprint arXiv:2307.09288}.

\bibitem[{Turpin et~al.(2023)Turpin, Michael, Perez, and
  Bowman}]{Turpin:2023aa}
Miles Turpin, Julian Michael, Ethan Perez, and Samuel~R. Bowman. 2023.
\newblock \href {http://arxiv.org/abs/2305.04388} {Language models don't always
  say what they think: Unfaithful explanations in chain-of-thought prompting}.

\bibitem[{Vafa et~al.(2021)Vafa, Deng, Blei, and
  Rush}]{vafa-etal-2021-rationales}
Keyon Vafa, Yuntian Deng, David Blei, and Alexander Rush. 2021.
\newblock \href {https://doi.org/10.18653/v1/2021.emnlp-main.807} {Rationales
  for sequential predictions}.
\newblock In \emph{Proceedings of the 2021 Conference on Empirical Methods in
  Natural Language Processing}, pages 10314--10332, Online and Punta Cana,
  Dominican Republic. Association for Computational Linguistics.

\bibitem[{Wang and Zhou(2024)}]{Wang:2024aa}
Xuezhi Wang and Denny Zhou. 2024.
\newblock \href {http://arxiv.org/abs/2402.10200} {Chain-of-thought reasoning
  without prompting}.

\bibitem[{Wang et~al.(2023)Wang, Kordi, Mishra, Liu, Smith, Khashabi, and
  Hajishirzi}]{wang-etal-2023-self-instruct}
Yizhong Wang, Yeganeh Kordi, Swaroop Mishra, Alisa Liu, Noah~A. Smith, Daniel
  Khashabi, and Hannaneh Hajishirzi. 2023.
\newblock \href {https://aclanthology.org/2023.acl-long.754} {Self-instruct:
  Aligning language models with self-generated instructions}.
\newblock In \emph{Proceedings of the 61st Annual Meeting of the Association
  for Computational Linguistics (Volume 1: Long Papers)}, pages 13484--13508,
  Toronto, Canada. Association for Computational Linguistics.

\bibitem[{Wei et~al.(2023)Wei, Wang, Schuurmans, Bosma, Ichter, Xia, Chi, Le,
  and Zhou}]{wei2023chainofthought}
Jason Wei, Xuezhi Wang, Dale Schuurmans, Maarten Bosma, Brian Ichter, Fei Xia,
  Ed~Chi, Quoc Le, and Denny Zhou. 2023.
\newblock \href {http://arxiv.org/abs/2201.11903} {Chain-of-thought prompting
  elicits reasoning in large language models}.

\bibitem[{Xue et~al.(2023)Xue, Fu, Zhou, Zheng, and
  You}]{paper:multiepoch-degradation}
Fuzhao Xue, Yao Fu, Wangchunshu Zhou, Zangwei Zheng, and Yang You. 2023.
\newblock To repeat or not to repeat: Insights from scaling llm under
  token-crisis.
\newblock \emph{arXiv preprint arXiv:2305.13230}.

\bibitem[{Yao et~al.(2023)Yao, Yu, Zhao, Shafran, Griffiths, Cao, and
  Narasimhan}]{paper:tree-of-thought}
Shunyu Yao, Dian Yu, Jeffrey Zhao, Izhak Shafran, Thomas~L Griffiths, Yuan Cao,
  and Karthik Narasimhan. 2023.
\newblock Tree of thoughts: Deliberate problem solving with large language
  models.
\newblock \emph{arXiv preprint arXiv:2305.10601}.

\bibitem[{Yao et~al.(2022)Yao, Zhao, Yu, Du, Shafran, Narasimhan, and
  Cao}]{paper:yao2022react}
Shunyu Yao, Jeffrey Zhao, Dian Yu, Nan Du, Izhak Shafran, Karthik Narasimhan,
  and Yuan Cao. 2022.
\newblock React: Synergizing reasoning and acting in language models.
\newblock \emph{arXiv preprint arXiv:2210.03629}.

\bibitem[{Ye et~al.(2023)Ye, Iyer, Celikyilmaz, Stoyanov, Durrett, and
  Pasunuru}]{ye2023complementary}
Xi~Ye, Srinivasan Iyer, Asli Celikyilmaz, Ves Stoyanov, Greg Durrett, and
  Ramakanth Pasunuru. 2023.
\newblock \href {http://arxiv.org/abs/2211.13892} {Complementary explanations
  for effective in-context learning}.

\bibitem[{Yuan et~al.(2023)Yuan, Yuan, Tan, Wang, Huang, and
  Huang}]{yuan2023rrhf}
Zheng Yuan, Hongyi Yuan, Chuanqi Tan, Wei Wang, Songfang Huang, and Fei Huang.
  2023.
\newblock \href {http://arxiv.org/abs/2304.05302} {{RRHF}: Rank responses to
  align language models with human feedback without tears}.

\bibitem[{Zheng et~al.(2023)Zheng, Mishra, Chen, Cheng, Chi, Le, and
  Zhou}]{zheng2023step}
Huaixiu~Steven Zheng, Swaroop Mishra, Xinyun Chen, Heng-Tze Cheng, Ed~H. Chi,
  Quoc~V Le, and Denny Zhou. 2023.
\newblock \href {http://arxiv.org/abs/2310.06117} {Take a step back: Evoking
  reasoning via abstraction in large language models}.

\bibitem[{Zhu et~al.(2023)Zhu, Jiao, and Jordan}]{zhu2023principled}
Banghua Zhu, Jiantao Jiao, and Michael~I. Jordan. 2023.
\newblock Principled reinforcement learning with human feedback from pairwise
  or $k$-wise comparisons.
\newblock \emph{arXiv preprint 2301.11270}.

\bibitem[{Ziegler et~al.(2020)Ziegler, Stiennon, Wu, Brown, Radford, Amodei,
  Christiano, and Irving}]{ziegler2020finetuning}
Daniel~M. Ziegler, Nisan Stiennon, Jeffrey Wu, Tom~B. Brown, Alec Radford,
  Dario Amodei, Paul Christiano, and Geoffrey Irving. 2020.
\newblock Fine-tuning language models from human preferences.
\newblock \emph{arXiv preprint 1909.08593}.

\end{thebibliography}
\bibliographystyle{acl_natbib}

\end{document}